\pdfoutput=1

\documentclass[11pt]{article}

\usepackage{acl}

\usepackage{times}
\usepackage{latexsym}

\usepackage[T1]{fontenc}

\usepackage[utf8]{inputenc}
\usepackage{microtype}

\title{Multilingual Generative Language Models for Zero-Shot \\ Cross-Lingual Event Argument Extraction}

\author{Kuan-Hao Huang\thanks{\; The authors contribute equally.}$^{\;\;\dagger}$ \ \ \ I-Hung Hsu\footnotemark[1]$^{\;\;\ddagger}$ \ \ \ Premkumar Natarajan$^{\ddagger}$ \\ {\bf Kai-Wei Chang$^{\dagger}$ \ \ \ Nanyun Peng$^{\dagger\ddagger}$}\\
$^{\dagger}$Computer Science Department, University of California, Los Angeles \\ 
$^{\ddagger}$Information Science Institute, University of Southern California \\
\texttt{\{khhuang, kwchang, violetpeng\}@cs.ucla.edu} \\
\texttt{\{ihunghsu, pnataraj\}@isi.edu} \\}

\usepackage{graphicx}
\usepackage{multirow}
\usepackage{makecell}
\usepackage{booktabs}
\usepackage{enumitem}
\usepackage{amssymb}
\usepackage{xcolor}
\usepackage{soul}

\usepackage{CJKutf8} 
\usepackage{amsmath}
\setuldepth{Berlin}
\usepackage{tabularx}
\newcolumntype{x}[1]{>{\arraybackslash\hspace{0pt}}m{#1}}

\newcommand{\mbf}[1]{\mathbf{#1}}

\newcommand{\model}[1]{\textsc{X-Gear}}

\usepackage{todonotes}

\begin{document}
\maketitle
\begin{abstract}
We present a study on leveraging multilingual pre-trained \emph{generative} language models for zero-shot cross-lingual event argument extraction (EAE).
By formulating EAE as a \emph{language generation} task, our method effectively encodes event structures and captures the dependencies between arguments. 
We design \emph{language-agnostic templates} to represent the event argument structures, which are compatible with any language, hence facilitating the cross-lingual transfer.
Our proposed model finetunes multilingual pre-trained generative language models to \emph{generate} sentences that fill in the language-agnostic template with arguments extracted from the input passage.
The model is trained on source languages and is then directly applied to target languages for event argument extraction.
Experiments demonstrate that the proposed model outperforms the current state-of-the-art models on zero-shot cross-lingual EAE. 
Comprehensive studies and error analyses are presented to better understand the advantages and the current limitations of using generative language models for zero-shot cross-lingual transfer EAE. 

\end{abstract}

\section{Introduction}
\label{sec:intro}

\begin{figure}[t!]
    \centering
    \includegraphics[trim=0cm 0cm 0cm 0cm, clip, width=0.95\columnwidth]{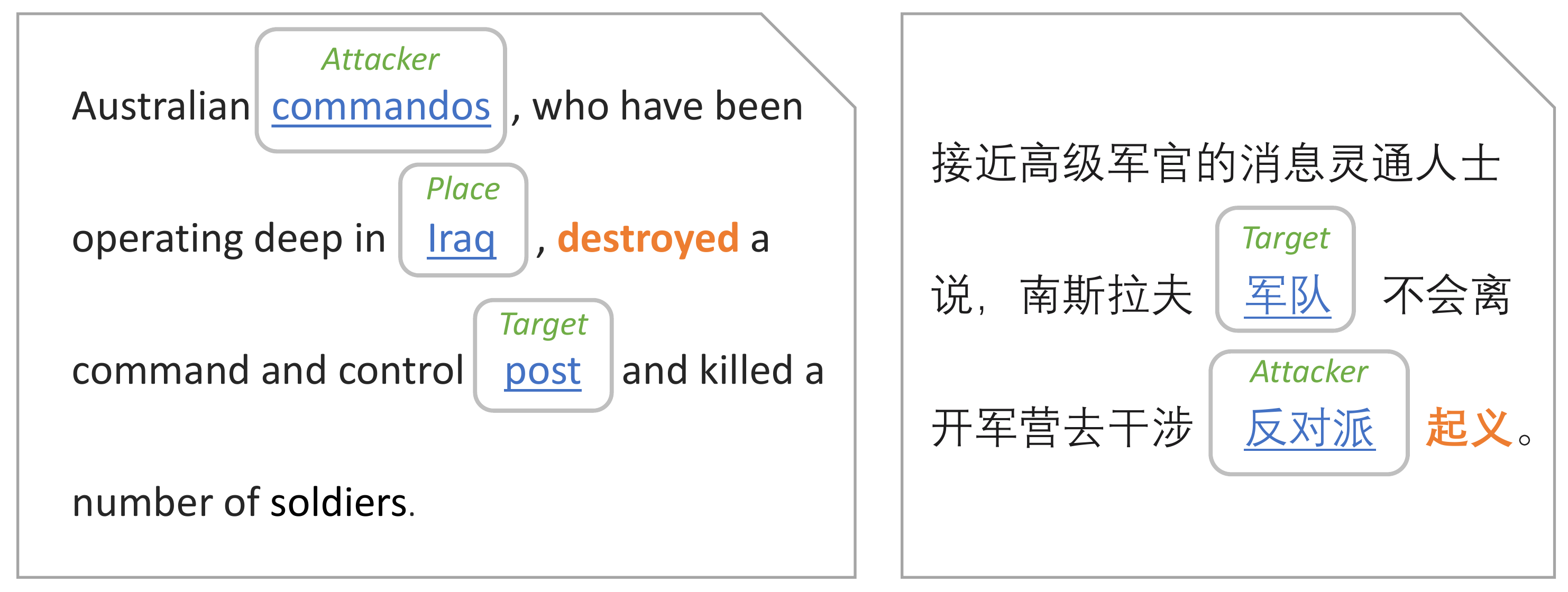}
    \caption{An illustration of cross-lingual event argument extraction. Given sentences in arbitrary languages and their event triggers (\emph{destroyed} and 起义), the model needs to identify arguments (\emph{commando}, \emph{Iraq} and \emph{post} v.s. 军队, and 反对派) and their corresponding roles (Attacker, Target, and Place).
    }
    \label{fig:intro_eae}
    \vspace{-1.2em}
\end{figure}

Event argument extraction (EAE) aims to recognize the entities serving as event arguments and identify their corresponding roles.
As illustrated by the English example in Figure~\ref{fig:intro_eae}, given a trigger word \emph{``destroyed''} for a \emph{Conflict:Attack} event, an event argument extractor is expected to identify \emph{``commando''}, \emph{``Iraq''}, and \emph{``post''} as the event arguments and predict their roles as \emph{``Attacker''},~\emph{``Place''}, and \emph{``Target''}, respectively.

Zero-shot cross-lingual EAE has attracted considerable attention since it eliminates the requirement of labeled data for constructing EAE models in low-resource languages \cite{Subburathinam19clgcn,Wasi21gate,Nguyen21unigcn}. In this setting, the model is trained on the examples in the \emph{source} languages and directly tested on the instances in the \emph{target} languages. 

\begin{CJK*}{UTF8}{gbsn}
\end{CJK*}

Recently, generation-based models\footnote{We use pre-trained \emph{generative} language models to refer to pre-trained models with encoder-decoder structure, such as BART~\cite{BART}, T5~\cite{t5model}, and mBART~\cite{Liu20mbart}. For models adapting these pre-trained generative models to generate texts for downstream applications, we denote them as \emph{generation-based} models.} have shown strong performances on monolingual structured prediction tasks \cite{Yan20genner,Huang21docgen,Paolini21tanl}, including EAE~\cite{Li21bartgen,Hsu21genee}. 
These works 
fine-tune pre-trained generative language models to generate outputs following designed templates such that the final predictions can be easily decoded from the outputs.
Compared to the traditional classification-based models \cite{Wang19hmeae,Wadden19dygiepp,Lin20oneie}, they better capture the structures and dependencies between entities, as the templates provide additional declarative information.

\begin{CJK*}{UTF8}{gbsn}
Despite the successes, the designs of templates in prior works are language-dependent, 
which makes it hard to be extended to the zero-shot cross-lingual transfer setting \cite{Subburathinam19clgcn,Wasi21gate}.
Naively applying such models trained on the source languages to the target languages usually generates \emph{code-switching} outputs, yielding poor performance for zero-shot cross-lingual transfer,\footnote{\label{note1}For example, TANL~\cite{Paolini21tanl} is trained to generate ``\texttt{[Two soldiers|target] were attacked}'' to represent \textit{Two soldiers} being a \textit{target} argument. When directly applying it to Chinese, the ground truth for TANL becomes ``\texttt{[两位士兵|target]被攻击}'', which is a sentence alternating between Chinese and English.} as we will empirically show in Section~\ref{sec:exp}.
How to design \emph{language-agnostic} generation-based models for zero-shot cross-lingual structured prediction problems is still an open question.
\end{CJK*}

In this work, we present a study that leverage multilingual pre-trained generative models for zero-shot cross-lingual event argument extraction and propose \model{} (\textbf{Cross}-lingual \textbf{G}enerative \textbf{E}vent \textbf{A}rgument extracto\textbf{R}). 
Given an input passage and a carefully designed prompt that contains an event trigger and the corresponding language-agnostic template, \model{} is trained to generate a sentence that fills in a language-agnostic template with arguments.
\model{} inherits the strength of generation-based models that captures event structures and the dependencies between entities better than classification-based models. Moreover, the pre-trained decoder inherently identifies named entities as candidates for event arguments and does not need an additional named entity recognition module. The \emph{language-agnostic templates} prevents the model from overfitting to the source language's vocabulary and facilitates cross-lingual transfer.

We conduct experiments on two multilingual EAE datasets: ACE-2005 \cite{Doddington04ace} and ERE \cite{Song15ere}. The results demonstrate that \model{} outperforms the state-of-the-art zero-shot cross-lingual EAE models. We further perform ablation studies to justify our design and present comprehensive error analyses to understand the limitations of using multilingual generation-based models for zero-shot cross-lingual transfer.  Our code is available at \url{https://github.com/PlusLabNLP/X-Gear}

\section{Related Work}

\paragraph{Zero-shot cross-lingual structured prediction.}
Zero-shot cross-lingual learning is an emerging research topic as it eliminates the requirement of labeled data for training models in low-resource languages \cite{Ruder21xtremer,Huang21robustxlt}. Various structured prediction tasks have been studied, including named entity recognition \cite{Pan17xner, huang2019matters,DBLP:conf/icml/HuRSNFJ20}, dependency parsing \cite{Ahmad19difficult,ahmad2019cross,meng2019target}, relation extraction \cite{Zou18re2,Ni19xre1}, and event argument extraction \cite{Subburathinam19clgcn,Nguyen21unigcn,DBLP:journals/corr/abs-2109-12383}.
Most of them are \emph{classification-based models} that build classifiers on top of a multilingual pre-trained \emph{masked} language models. 
To further deal with the discrepancy between languages, some of them require additional information, such as bilingual dictionaries \cite{Liu19cross-align,Ni19xre1}, translation pairs \cite{Zou18re2}, and dependency parse trees \cite{Subburathinam19clgcn,Wasi21gate,Nguyen21unigcn}.
However, as pointed out by previous literature \cite{Li21bartgen,Hsu21genee}, classification-based models are less powerful to model dependencies between entities compared to \emph{generation-based models}.

\paragraph{Generation-based structured prediction.}
Several works have demonstrated the great success of generation-based models on monolingual structured prediction tasks, including named entity recognition \cite{Yan20genner}, relation extraction \cite{Huang21docgen,Paolini21tanl}, and event extraction  \cite{Du21grit,Li21bartgen,Hsu21genee,Lu21text2event}. Yet, as mentioned in Section~\ref{sec:intro}, their designed generating targets are language-dependent. Accordingly, directly applying their methods to the zero-shot cross-lingual setting would result in less-preferred performance.

\paragraph{Prompting methods.}
There are growing interests recently to incorporate prompts on pre-trained language models in order to guide the models' behavior or elicit knowledge~\cite{DBLP:conf/naacl/PengPFD019,sheng2020towards,DBLP:conf/emnlp/ShinRLWS20,DBLP:conf/eacl/SchickS21,QinE21,ScaoR21}.
Following the taxonomy in~\cite{Liu21promptsurvey}, these methods can be classified depending on whether the language models' parameters are tuned and on whether trainable prompts are introduced. Our method belongs to the category that fixes the prompts and tunes the language models' parameters. Despite the flourish of the research in prompting methods, there is only limited attention being put on multilingual tasks~\cite{DBLP:journals/corr/abs-2109-07684}.

\begin{figure*}[t!]
    \centering
    \includegraphics[width=0.98\textwidth]{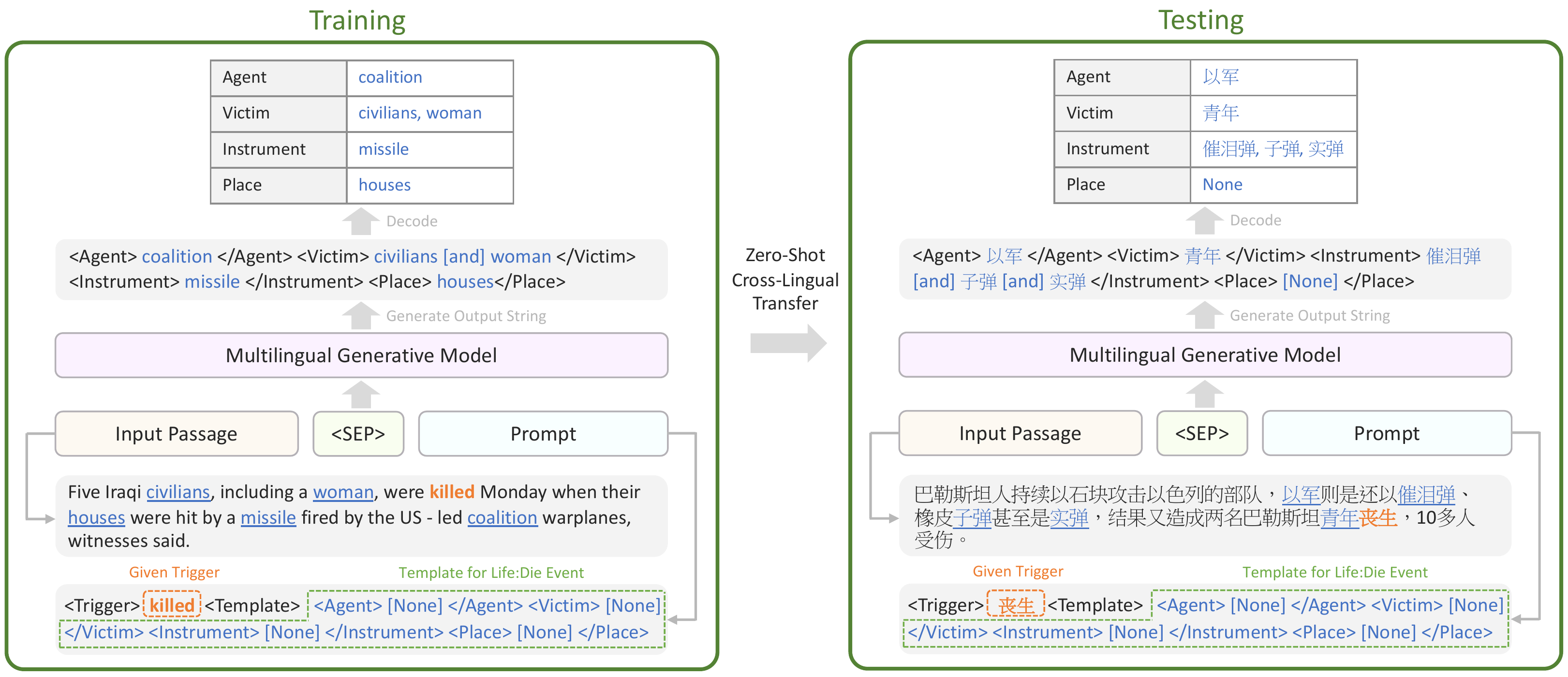}
    \caption{The overview of \model{}. Given an input passage and a carefully designed prompt containing an event trigger and a language-agnostic template, \model{} fills in the language-agnostic template with event arguments.}
    \label{fig:overview}
    \vspace{-0.8em}
\end{figure*}

\section{Zero-Shot Cross-Lingual Event Argument Extraction}
We focus on zero-shot cross-lingual EAE. 
Given an input passage and an event trigger, an EAE model identifies arguments and their corresponding roles.
More specifically, as illustrated by the training examples in Figure~\ref{fig:overview}, given an input passage $\mbf x$ and an event trigger $\mbf t$ (\emph{killed}) belonging to an event type $\mbf e$ (\emph{Life:Die}), an EAE model  predicts a list of arguments $\mbf a = [a_1, a_2, ..., a_l]$ (\emph{coalition}, \emph{civilians}, \emph{woman}, \emph{missile}, \emph{houses}) and their corresponding roles $\mbf r = [r_1, r_2, .., r_l]$ (\emph{Agent}, \emph{Victim}, \emph{Victim}, \emph{Instrument}, \emph{Place}). 
In the zero-shot cross-lingual setting, the training set $X_{train}$ = $\{(\mbf x_i, \mbf t_i, \mbf e_i, \mbf a_i,\mbf r_i)\}_{i=1}^N$ belongs to the source languages while the testing set $X_{test}$ = $\{(\mbf x_i, \mbf t_i, \mbf e_i, \mbf a_i,\mbf r_i)\}_{i=1}^M$ are in the target languages.

Similar to monolingual EAE, zero-shot cross-lingual EAE models are expected to capture the dependencies between arguments and make structured predictions. 
However, unlike monolingual EAE, zero-shot cross-lingual EAE models need to handle the differences (e.g., grammar, word order) between languages and learn to transfer the knowledge from the source languages to the target languages.
\section{Proposed Method: \model{}}
\label{sec:method}
We formulate zero-shot cross-lingual EAE as a language generation task and propose \model{}, a \textbf{Cross}-lingual \textbf{G}enerative \textbf{E}vent \textbf{A}rgument extracto\textbf{R} that is illustrated in Figure~\ref{fig:overview}.
There are two challenges raised by this formulation: (1) The input language may vary during training and testing; (2) The generated output strings need to be easily parsed into final predictions. Therefore, the output strings have to reflect the change of the input language accordingly while remaining well-structured. 

We address these challenges by designing \emph{language-agnostic templates}. Specifically, given an input passage $\mbf x$ and a designed prompt that contains the given trigger $\mbf t$, its event type $\mbf e$, and a \emph{language-agnostic template}, \model{} learns to generate an output string that fills in the language-agnostic template with information extracted from input passage.
The language-agnostic template is designed in a structured way such that parsing the final argument predictions~$\mbf a$ and role predictions~$\mbf r$ from the generated output is trivial. 
Moreover, since the template is language-agnostic, it facilitates cross-lingual transfer. 

\model{} fine-tunes multilingual pre-trained generative models, such as mBART-50 \cite{mbart50} or mT5 \cite{Xue21mt5}, and augments them with a copy mechanism to better adapt to input language changes.
We present its details as follows, including the language-agnostic templates, the target output string, the input format, and the training details.

\subsection{Language-Agnostic Template} 

We create one language-agnostic template $T_{\mbf e}$ for each event type $\mbf e$, in which we list all possible associated roles\footnote{The associated roles can be obtained by skimming training data or directly from the annotation guideline if provided.} and form a unique HTML-tag-style template for that event type $\mbf e$. For example, in Figure~\ref{fig:overview}, the \emph{Life:Die} event is associated with four roles: \emph{Agent}, \emph{Victim}, \emph{Instrument}, and \emph{Place}. Thus, the template for \emph{Life:Die} events is designed as:
\\[6pt]
{\setlength{\tabcolsep}{1pt}
\begin{tabular}{|x{0.99\columnwidth}|}
    \hline
    \fontsize{7.8pt}{7.8pt}\selectfont
    $<$Agent$>$\texttt{[None]}$<$/Agent$>$$<$Victim$>$\texttt{[None]}$<$/Victim$>$ \\
    \fontsize{7.8pt}{7.8pt}\selectfont
    $<$Instrument$>$\texttt{[None]}$<$/Instrument$>$$<$Place$>$\texttt{[None]}$<$/Place$>$. \\
    \hline
\end{tabular}}
\\

For ease of understanding, we use English words to present the template. However, these tokens (\texttt{[None]}, <Agent>, </Agent>, <Victim>, etc.) are encoded as special tokens\footnote{In fact , the special tokens can be replaced by any other format, such as <--token1--> or </--token1-->. Here, we use <Agent> and </Agent> to highlight that arguments between these two special tokens are corresponding to the \textit{Agent} role.} that the pre-trained models have never seen and thus their representations need to be learned from scratch. Since these special tokens are not associated with any language and are not pre-trained, they are considered as \emph{language-agnostic}.

\subsection{Target Output String}
\model{} learns to generate target output strings that follow the form of language-agnostic templates.
To compose the target output string for training, given an instance $(\mbf x, \mbf t, \mbf e, \mbf a, \mbf r)$, 
we first pick out the language-agnostic template $T_{\mbf e}$ for the event type $\mbf e$ and then replace all ``\texttt{[None]}'' in $T_{\mbf e}$ with the corresponding arguments in $\mbf a$ according to their roles $\mbf r$. If there are multiple arguments for one role, we concatenate them with a special token ``\texttt{[and]}''. 
For instance, the training example in Figure~\ref{fig:overview} has two arguments (\emph{civilians} and \emph{woman}) for the \emph{Victim} role, and the corresponding part of the output string would be
\\[6pt]
{\setlength{\tabcolsep}{2pt}
\begin{tabular}{|x{0.99\columnwidth}|}
    \hline
    \fontsize{7.8pt}{7.8pt}\selectfont
    $<$Victim$>$ \texttt{civilians [and] woman} $<$/Victim$>$. \\
    \hline
\end{tabular}}
\\[6pt]
If there are no corresponding arguments for one role, we keep ``\texttt{[None]}'' in $T_{\mbf e}$.
By applying this rule, the full output string for the training example in Figure~\ref{fig:overview} becomes
\\[6pt]
{\setlength{\tabcolsep}{2pt}
\begin{tabular}{|x{0.99\columnwidth}|}
    \hline
    \fontsize{7.8pt}{7.8pt}\selectfont
    $<$Agent$>$ \texttt{coalition} $<$/Agent$>$$<$Victim$>$ \texttt{civilians[and]} \\
    \fontsize{7.8pt}{7.8pt}\selectfont
    \texttt{woman} $<$/Victim$>$$<$Instrument$>$ \texttt{missile} $<$/Instrument$>$\\ 
    \fontsize{7.8pt}{7.8pt}\selectfont
    $<$Place$>$ \texttt{houses} $<$/Place$>$. \\
    \hline
\end{tabular}}
\\

Since the output string is in the HTML-tag style, we can easily decode the argument and role predictions from the generated output string via a simple rule-based algorithm.

\subsection{Input Format}
As we mentioned previously, the key for the generative formulation for zero-shot cross-lingual EAE is to guide the model to generate output strings in the desired format.
To facilitate this behavior, we feed the input passage $\mbf x$ as well as a \emph{prompt} to \model{}, as shown by Figure~\ref{fig:overview}.
The \emph{prompt} contains all valuable information for the model to make predictions, including a trigger $\mbf t$ and a language-agnostic template $T_{\mbf e}$. Notice that we do not \textit{explicitly} include the event type $\mbf e$ in the prompt because the template $T_{\mbf e}$ \textit{implicitly} contains this information.
In Section~\ref{sec:ablation}, we will show the experiments on explicitly adding event type $\mbf e$ to the prompt and discuss its influence on the cross-lingual transfer.

\subsection{Training}
To enable \model{} to generate sentences in different languages, we resort multilingual pre-trained generative model to be our base model, which models the conditional probability of generating a new token given the previous generated tokens and the input context to the encoder $c$, i.e,
\vspace{-0.1em}
\[ P(x|c)=\prod_{i} P_{gen}(x_i|x_{<i},c), \vspace{-0.3em}\]
where $x_i$ is the output of the decoder at step $i$. 

\paragraph{Copy mechanism.}
Although the multilingual pre-trained generative models can generate sequences in many languages, solely relying on them may result in generating hallucinating arguments~\cite{Li21bartgen}. Since most of the tokens in the target output string appear in the input sequence,\footnote{Except for the special tokens \texttt{[and]} and \texttt{[None]}.} we augment the multilingual pre-trained generative models with a copy mechanism to help \model{} better adapt to the cross-lingual scenario.
Specifically, we follow \citet{See17copy} to decide the conditional probability of generating a token $t$ as a weighted sum of the vocabulary distribution computed by multilingual pre-trained generative model $P_{gen}$ and copy distribution $P_{copy}$
\vspace{-0.2em}
\begin{equation*}
\small
\begin{split}
P_{\model{}}(x_i=t|x_{<i},c)& = \\
    w_{copy} \cdot P_{copy}(t) + &(1-w_{copy}) \cdot P_{gen}(x_i=t|x_{<i},c)
\end{split}
\end{equation*}
where $w_{copy} \in [0,1]$ is the copy probability computed by passing the decoder hidden state at time step $i$ to a linear layer. As for $P_{copy}$, it refers to the probability over input tokens weighted by the cross-attention that the last decoder layer computed (at time step $i$). Our model is then trained end-to-end with the following loss:
\vspace{-0.3em}
\[ \mathcal{L}=-\log \sum_{i} P_{\model{}}(x_i|x_{<i},c). \vspace{-0.3em}\]
\section{Experiments}

\subsection{Datasets}
We consider two commonly used event extraction datasets: ACE-2005 and ERE. We consider English, Arabic, and Chinese annotations for \textbf{ACE-2005} \cite{Doddington04ace} and 
follow the preprocessing in \citet{Wadden19dygiepp} to keep 33 event types and 22 argument roles. 
\textbf{ERE} \cite{Song15ere} is created by the Deep Exploration and Filtering of Test program. We consider its English and Spanish annotations and follow the preprocessing in \citet{Lin20oneie} to keep 38 event types and 21 argument roles. Detailed statistics and preprocessing steps about the two datasets are in Appendix~\ref{app:dataset}.

Notice that prior works working on the zero-shot cross-lingual transfer of event arguments mostly focus on event argument role labeling~\cite{Subburathinam19clgcn,Wasi21gate}, where they assume ground truth entities are provided during both training and testing. In their experimental data splits, events in a sentence can be scattered in all training, development, and test split since they treat each event-entity pair as a different instance. In this work, we consider event argument extraction \cite{Wang19hmeae,Wadden19dygiepp,Lin20oneie}, which is a more realistic setting.

\subsection{Evaluation Metric}
We follow previous work \cite{Lin20oneie,Wasi21gate} and consider the \emph{argument classification F1 score} to measure the performance of models. An argument-role pair is counted as correct if both the argument offsets and the role type match the ground truth. Given the ground truth arguments $\mbf a$, ground truth roles $\mbf r$, predicted arguments $\tilde{\mbf a}$, and predicted roles $\tilde{\mbf r}$, the argument classification F1 score is defined as the F1 score between the set $\{ (\mbf a_i, \mbf r_i)\}$ and the set $\{ (\tilde{\mbf a}_j, \tilde{\mbf r}_j)\}$. 
For every model, we experiment with three different random seeds and report the average results.

\begin{table*}[t!]
\centering
\small
\setlength{\tabcolsep}{5pt}
\resizebox{\textwidth}{!}{
\begin{tabular}{l|c|ccc|ccc|ccc|c}
    \toprule
    Model & \makecell{\# of \\ parameters}
    &\makecell{en\\$\Downarrow$\\en} & \makecell{en\\$\Downarrow$\\zh} & \makecell{en\\$\Downarrow$\\ar}
    &\makecell{ar\\$\Downarrow$\\ar} & \makecell{ar\\$\Downarrow$\\en} & \makecell{ar\\$\Downarrow$\\zh}
    &\makecell{zh\\$\Downarrow$\\zh} & \makecell{zh\\$\Downarrow$\\en} & \makecell{zh\\$\Downarrow$\\ar}
    & avg \\
    \midrule
    OneIE (XLM-R-large) \cite{Lin20oneie} & $\sim$570M
    & 63.6 & 42.5 & 37.5 & 57.8 & 27.5 & \ul{31.2} & \ul{69.6} & 51.5 & 31.1 & 45.8 \\
    CL-GCN (XLM-R-large) \cite{Subburathinam19clgcn} & $\sim$570M 
    & 59.8 & 29.4 & 25.0 & 47.5 & 25.4 & 19.4 & 62.2 & 40.8 & 23.3 & 37.0 \\
    GATE (XLM-R-large) \cite{Wasi21gate} & $\sim$590M
    & 67.0 & 49.2 & \ul{44.5} & 59.6 & 27.6 & 26.3 & \textbf{70.6} & 46.7 & \textbf{37.3} & 47.6 \\
    GATE (mBART-50-large) & $\sim$630M
    & 65.5 & 43.0 & 38.9 & 58.5 & 27.5 & 26.1 & 65.9 & 45.3 & 30.2 & 44.5 \\
    GATE (mT5-base) & $\sim$590M
    & 59.8 & 47.7 & 32.6 & 45.4 & 20.7 & 21.0 & 64.0 & 35.3 & 22.8 & 38.8 \\
    \midrule
    TANL (mT5-base) \cite{Paolini21tanl} & $\sim$580M
    & 59.1 & 38.6 & 29.7 & 50.1 & 18.3 & 16.9 & 65.2 & 33.3 & 18.3 & 36.6 \\
    \midrule
    \model{} (mBART-50-large) & $\sim$610M 
    & \ul{68.3} & 48.9 & 37.8 & 59.8 & \ul{30.5} & 29.2 & 63.6 & 45.9 & 32.3 & 46.2 \\
    \model{} (mT5-base) & $\sim$580M 
    & 67.9 & \ul{53.1} & 42.0 & \ul{66.2} & 27.6 & 30.5 & 69.4 & \ul{52.8} & 32.0 & \ul{49.1} \\
    \midrule
    \model{} (mT5-large) & $\sim$1230M 
    & \textbf{71.2} & \textbf{54.0} & \textbf{44.8} & \textbf{68.9} & \textbf{32.1} & \textbf{33.3} & 68.9 & \textbf{55.8} & \ul{33.1} & \textbf{51.3} \\
    \bottomrule

\end{tabular}}
\caption{Average results in argument classification F1(\%) of ACE-2005 with three different seeds. The best is in bold and the second best is underlined. ``en $\Rightarrow$ zh'' denotes models transferring from en to zh. Compared with models using similar numbers of parameters, \model{} (mT5-base) outperforms baselines. To test the influence of using larger pre-trained generative models, we add \model{} (mT5-large), which achieves even better results.}
\vspace{-0.8em}
\label{tab:ace}
\end{table*}

\begin{table}[t!]
\centering
\small
\setlength{\tabcolsep}{4.7pt}
\resizebox{.47\textwidth}{!}{
\begin{tabular}{l|cc|cc|c}
    \toprule
    Model 
    &\makecell{en\\$\Downarrow$\\en} & \makecell{en\\$\Downarrow$\\es} 
    &\makecell{es\\$\Downarrow$\\es} & \makecell{es\\$\Downarrow$\\en} 
    & avg \\
    \midrule
    OneIE (XLM-R-large)
    & 64.4 & 56.8 & 64.8 & 56.9 & 60.7  \\
    CL-GCN (XLM-R-large) 
    & 61.9 & 51.9 & 62.9 & 48.5 & 55.9  \\
    GATE (XLM-R-large) 
    & 66.4 & \textbf{61.5} & 63.0 & 56.5 & 61.9  \\
    \midrule
    TANL (mT5-base)
    & 65.9 & 40.3 & 58.6 & 47.4 & 53.1  \\
    \midrule
    \model{} (mBART-50-large)
    & 69.5 & 57.3 & 63.9 & 58.9 & 62.4  \\
    \model{} (mT5-base)
    & \ul{69.8} & 57.9 & \ul{66.1} & \ul{59.0} & \ul{63.2}  \\
    \midrule
    \model{} (mT5-large)
    & \textbf{72.9} & \ul{59.7} & \textbf{67.4} & \textbf{64.1} & \textbf{66.0}  \\
    \bottomrule

\end{tabular}}
\caption{Average results in argument classification F1(\%) of ERE with three different seeds. The best is in bold and the second best is underlined. ``en $\Rightarrow$ es'' denotes that models transfer from en to es.}
\vspace{-1.2em}
\label{tab:ere}
\end{table}

\subsection{Compared Models}
We compare the following models and their implementation details are listed in Appendix~\ref{app:imp}.
\begin{itemize}[topsep=3pt, itemsep=-1pt, leftmargin=12pt]
    \item \textbf{OneIE} \cite{Lin20oneie}, the state-of-the-art for monolingual event extraction, is a classification-based model trained with multitasking, including entity extraction, relation extraction, event extraction, and \emph{event argument extraction}. We simply replace its pre-trained embedding with XLM-RoBERTa-large \cite{Conneau20xlmr} to fit the zero-shot cross-lingual setting. Note that the multi-task learning makes OneIE require \emph{additional annotations}, such as named entity annotations and relation annotations.
    \item \textbf{CL-GCN} \cite{Subburathinam19clgcn} 
    is a classification-based model for cross-lingual event argument role labeling (EARL). It considers \emph{dependency parsing annotations} to bridge different languages and use GCN layers \cite{Kipf17gcn} to encode the parsing information. We follow the implementation of previous work \cite{Wasi21gate} and add two GCN layers on top of XLM-RoBERTa-large. Since CL-GCN focuses on EARL tasks, which assume the ground truth entities are available during testing, we add one name entity recognition module jointly trained with CL-GCN. 
    \item \textbf{GATE} \cite{Wasi21gate}, the state-of-the-art model for zero-shot cross-lingual EARL, is a classification-based model which considers \emph{dependency parsing annotations} as well. Unlike CL-GCN, it uses a Transformer layer \cite{Vaswani17transformer} with modified attention to encode the parsing information. We follow the original implementation and add two GATE layers on top of pre-trained multilingual language models.\footnote{To better compare our method with this strong baseline, we consider three different pre-trained multilingual language models for GATE -- (1) XLM-RoBERTa-large (2) mBART-50-large (3) mT5-base. For mBART-50-large and mT-base, we follow BART's recipe~\cite{BART} to extract features for EAE predictions. Specifically, the input passage is fed into both encoder and decoder, and the final token representations are elicited from the decoder output.} Similar to CL-GCN, we add one name entity recognition module jointly trained with GATE. 
    \item \textbf{TANL} \cite{Paolini21tanl} is a generation-based model for monolingual EAE. Their predicted target is a sentence that embeds labels into the input passage, such as \texttt{[Two soldiers|target] were attacked}, which indicates that ``\textit{Two soldiers}'' is a ``\textit{target}'' argument. To adapt TANL to zero-shot cross-lingual EAE, we change its pre-trained generative model from T5 \cite{t5model} to mT5-base \cite{Xue21mt5}.
    \item \textbf{\model{}} is our proposed model. We consider three different pre-trained generative language models: mBART-50-large \cite{mbart50}, mT5-base, and mT5-large \cite{Xue21mt5}.
\end{itemize}

\subsection{Results}
\label{sec:exp}

Table~\ref{tab:ace} and Table~\ref{tab:ere} list the results on ACE-2005 and ERE, respectively,  with all combinations of source languages and target languages. Note that all the models have similar numbers of parameters except for \model{} with mT5-large.

\paragraph{Comparison to prior generative models.}
We first observe that TANL has poor performance when transferring to different languages. The reason is that its language-dependent template makes TANL easily generate code-switching outputs,\footnote{Such as the example shown in footnote~\ref{note1}.} which is a case that pre-trained generative model rarely seen, leading to poor performance.
In contrast, \model{} considers the language-agnostic templates and achieves better performance for zero-shot cross-lingual transfer.

\paragraph{Comparison to classification models.}
\model{} with mT5-base outperforms OneIE, CL-GCN, and GATE on almost all the combinations of the source language and the target language. This suggests that our proposed method is indeed a promising approach for zero-shot cross-lingual EAE.

It is worth noting that OneIE, CL-GCN, and GATE require an additional pipeline named entity recognition module to make predictions. Moreover, CL-GCN and GATE need additional dependency parsing annotations to align the representations of different languages. 
On the contrary, \model{} is able to leverage the learned knowledge from the pre-trained generative models, and therefore no additional modules or annotations are needed.

\paragraph{Comparison to different pre-trained generative language models.}
Interestingly, using mT5-base is more effective than using mBART-50-large for \model{}, although they have a similar amount of parameters. We conjecture that the use of special tokens leads to this difference. mBART-50 has different begin-of-sequence (BOS) tokens for different languages. During generation, we have to specify which BOS token we would like to use as the start token. We guess that this language-specific BOS token makes mBART-50 harder to transfer the knowledge from the source language to the target language. Unlike mBART-50, mT5 does not have such language-specific BOS tokens. During generation, mT5 uses the padding token as the start token to generate a sequence. This design is more general and benefit zero-shot cross-lingual transfer.

\paragraph{Larger pre-trained models are better.}
Finally, we demonstrate that the performance of \model{} can be further boosted with a larger pre-trained generative language model. As shown by Table~\ref{tab:ace} and Table\ref{tab:ere}, \model{} with mT5-large achieves the best scores on most of the cases.

\section{Analysis}

\subsection{Ablation Studies}
\label{sec:ablation}

\paragraph{Copy mechanism.}
We first study the effect of the copy mechanism. Table~\ref{tab:copy} lists the performance of \model{} with and without copy mechanism. 
It shows improvements in adding a copy mechanism when using mT5-large and mT-base.
However, interestingly, adding a copy mechanism is not effective for mBART-50. We conjecture that this is because the pre-trained objective of mBART-50 is denoising autoencoding \cite{Liu20mbart}, and it has already learned to copy tokens from the input. Therefore, adding a copy mechanism is less useful. In contrast, the pre-trained objective of mT5 is to only generate tokens been masked out, resulting in lacking the ability to copy input. Thus, the copy mechanism becomes beneficial for mT5.

\begin{table}[t!]
\centering
\small
\setlength{\tabcolsep}{4.5pt}
\resizebox{\columnwidth}{!}{
\begin{tabular}{l|ccc|ccc|c}
    \toprule
    Model
    &\makecell{en\\$\Downarrow$\\xx} & \makecell{ar\\$\Downarrow$\\xx} & \makecell{zh\\$\Downarrow$\\xx}
    &\makecell{xx\\$\Downarrow$\\en} & \makecell{xx\\$\Downarrow$\\ar} & \makecell{xx\\$\Downarrow$\\zh}
    & avg \\
    \midrule
    mBART-50-large
    & \textbf{51.6} & 39.8 & 47.2 & 48.2 & 43.2 & 47.2 & 46.2 \\
    \ - w/o copy
    & 50.9 & \textbf{42.2} & \textbf{49.6} & \textbf{50.6} & \textbf{43.5} & \textbf{48.7} & \textbf{47.6} \\
    \midrule
    mT5-base
    & \textbf{54.3} & \textbf{41.4} & \textbf{51.4} & \textbf{49.4} & \textbf{46.7} & \textbf{51.0} & \textbf{49.1} \\
    \ - w/o copy
    & 52.1 & 39.5 & 47.6 & 48.1 & 42.7 & 48.5 & 46.4 \\
    \midrule
    mT5-large
    & \textbf{56.7} & 44.8 & \textbf{52.6} & \textbf{53.0} & \textbf{48.9} & 52.1 & \textbf{51.3} \\
    \ - w/o copy
    & 55.1 & \textbf{45.0} & 51.5 & 52.0 & 46.3 & \textbf{53.2} & 50.5 \\
    \bottomrule

\end{tabular}}
\caption{Ablation study on copy mechanism for ACE-2005. ``en $\Rightarrow$ xx'' indicates the average of ``en $\Rightarrow$ en'', ``en $\Rightarrow$ zh'', and ``en $\Rightarrow$ ar''.}
\vspace{-1.3em}
\label{tab:copy}
\end{table}

\paragraph{Including event type in prompts.}
In Section~\ref{sec:method}, we mentioned that the designed prompt for \model{} consists of only the input sentence and the language-agnostic template. In this section, we discuss whether \textit{explicitly} including the event type information in the prompt is helpful. We consider three ways to include the event type information:
\begin{CJK*}{UTF8}{gbsn}
\begin{itemize}[topsep=3pt, itemsep=0.5pt, leftmargin=13pt]
    \item \textbf{English tokens}. We put the English version of the event type in the prompt even if we are training or testing on non-English languages, for example, using \textit{Attack} for the event type \textit{Attack}.
    \item \textbf{Translated tokens}. 
    For each event type, we prepare the translated version of that event type token. For example, both \textit{Attack} and \textit{攻击}\ represents the \textit{Attack} event type.
    During training or testing, we decide the used token(s) according to the language of the input passage. Since all the event types are written in English in ACE-2005 and ERE, we use an off-the-self machine translation tool to perform the translation.
    \item \textbf{Special tokens}. We create a special token for every event type and let the model learn the representations of the special tokens from scratch. For instance, we use \texttt{<--attack-->} to represent the \textit{Attack} event type.
\end{itemize}
\end{CJK*}

\begin{CJK*}{UTF8}{gbsn}
Table~\ref{tab:prompt} shows the results. In most cases, including event type information in the prompt decreases the performance. One reason is that one word in a language can be mapped to several words in another language. For example, the \emph{Life} event type is related to \emph{Marry}, \emph{Divorce}, \emph{Born}, and \emph{Die} four sub-event types. In English, we can use just one word \emph{Life} to cover all four sub-event types. However, In Chinese, when talking about \emph{Marry} and \emph{Divorce}, \emph{Life} should be translated to ``生活''; when talking about \emph{Born} and \emph{Die}, \emph{Life} should be translated to ``生命''. This mismatch may cause the performance drop when considering event types in prompts. 
We leave how to efficiently use event type information in the cross-lingual setting as future work. 
\end{CJK*}

\begin{table}[t!]
\centering
\small
\setlength{\tabcolsep}{3.5pt}
\resizebox{\columnwidth}{!}{
\begin{tabular}{@{}l|ccc|ccc|c}
    \toprule
    Model
    &\makecell{en\\$\Downarrow$\\xx} & \makecell{ar\\$\Downarrow$\\xx} & \makecell{zh\\$\Downarrow$\\xx}
    &\makecell{xx\\$\Downarrow$\\en} & \makecell{xx\\$\Downarrow$\\ar} & \makecell{xx\\$\Downarrow$\\zh}
    & avg \\
    \midrule
    \model{} (mT5-base)
    & \textbf{54.3} & \textbf{41.4} & 51.4 & 49.4 & \textbf{46.7} & \textbf{51.0} & \textbf{49.1} \\
    \ w/ English Tokens
    & 53.3 & 39.3 & \textbf{52.3} & 49.2 & 46.5 & 49.2 & 48.3 \\
    \ w/ Translated Tokens
    & 51.7 & 40.4 & 52.2 & \textbf{49.8} & 45.6 & 48.8 & 48.1 \\
    \ w/ Special Tokens
    & 52.3 & 39.7 & 51.8 & 49.0 & 45.4 & 49.3 & 47.9 \\
    \bottomrule
\end{tabular}}
\caption{Ablation study on including event type information in prompts for ACE-2005. ``en $\Rightarrow$ xx'' indicates the average of ``en $\Rightarrow$ en'', ``en $\Rightarrow$ zh'', and ``en $\Rightarrow$ ar''.}
\label{tab:prompt}
\vspace{-0.1em}
\end{table}

\begin{figure*}[t!]
    \centering
    \includegraphics[width=0.75\linewidth]{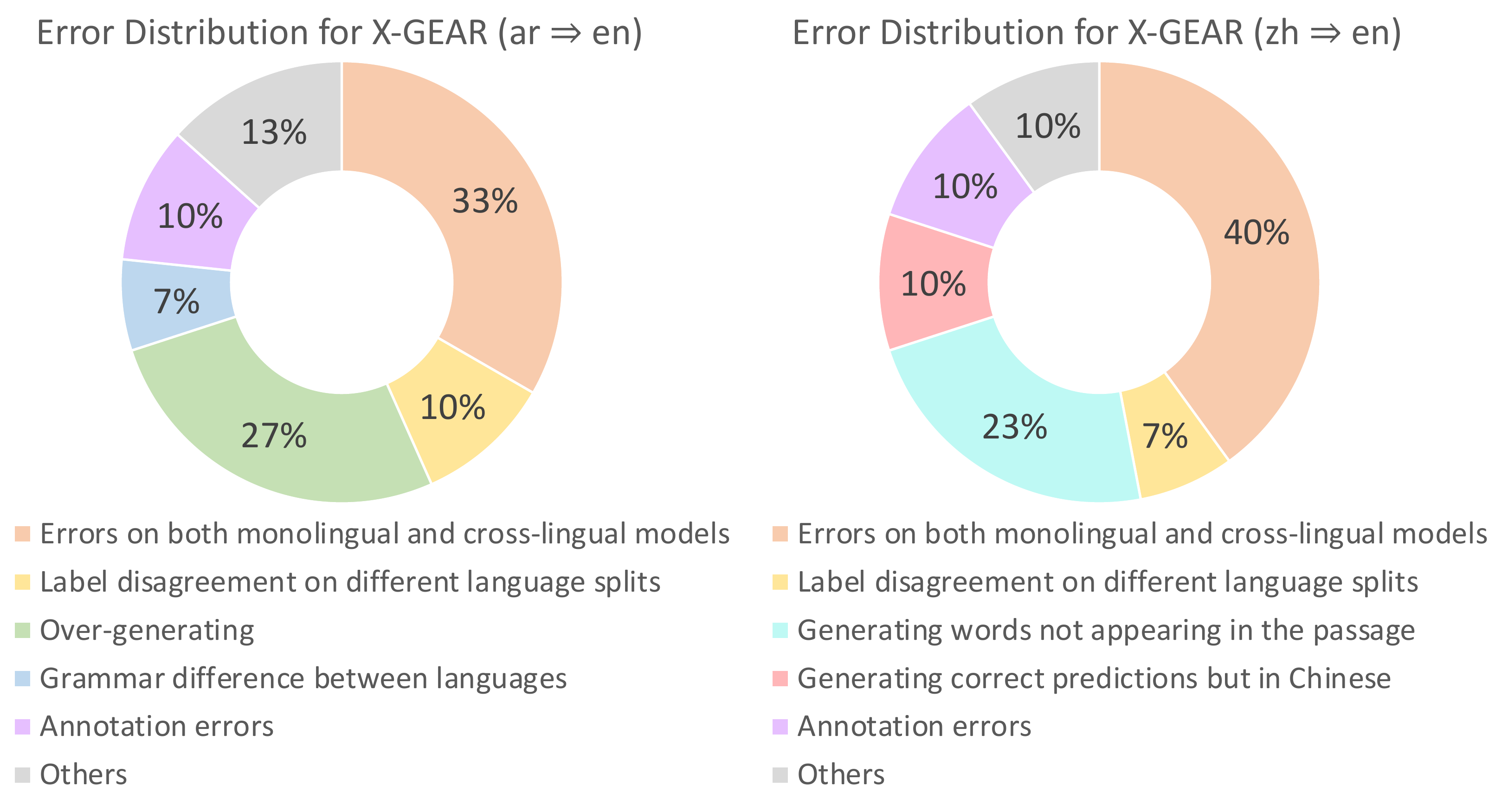}
    \caption{Distribution of errors that made by \model{} (mT5-base). \textbf{\textit{Left:}} The distribution for our model that transfers from Arabic to English; \textbf{\textit{Right:}} The distribution for our model trained on Chinese and tested on English.}
    \vspace{-0.3em}
    \label{fig:err_analysis}
\end{figure*}

\paragraph{Influence of role order in templates.}
The order of roles in the designed language-agnostic templates can potentially influence performance. When designing the templates, we intentionally make the order of roles close to the order in natural sentences.\footnote{For example, types related to subject and object are listed first and types related to methods and places are listed last.} To study the effect of different orders, we train \model{} with templates with different random orders and report the results in Table~\ref{tab:ablation}. \model{} with random orders still achieve good performance but slightly worse than the original order. It suggests that \model{} is not very sensitive to different templates while providing appropriate order of roles can lead to a small improvement.

\begin{table}[t!]
\centering
\small
\setlength{\tabcolsep}{3.5pt}
\resizebox{\columnwidth}{!}{
\begin{tabular}{@{}l|ccc|ccc|c}
    \toprule
    Model
    &\makecell{en\\$\Downarrow$\\xx} & \makecell{ar\\$\Downarrow$\\xx} & \makecell{zh\\$\Downarrow$\\xx}
    &\makecell{xx\\$\Downarrow$\\en} & \makecell{xx\\$\Downarrow$\\ar} & \makecell{xx\\$\Downarrow$\\zh}
    & avg \\
    \midrule
    \model{} (mT5-base)
    & 54.3 & \textbf{41.4} & \textbf{51.4} & 49.4 & \textbf{46.7} & \textbf{51.0} & \textbf{49.1} \\
    \ w/ random order 1
    & \textbf{54.4} & 38.9 & 50.8 & 48.7 & 45.1 & 50.1 & 48.0 \\
    \ w/ random order 2
    & 52.1 & 40.4 & \textbf{51.4} & 48.3 & 45.9 & 49.7 & 48.0 \\
    \ w/ random order 3
    & 53.7 & 40.8 & 50.7 & \textbf{50.8} & 45.8 & 48.6 & 48.4 \\
    \bottomrule
\end{tabular}}
\caption{Ablation study on different orders of roles in templates for ACE-2005. ``en $\Rightarrow$ xx'' indicates the average of ``en $\Rightarrow$ en'', ``en $\Rightarrow$ zh'', and ``en $\Rightarrow$ ar''.}
\label{tab:ablation}
\vspace{-1.3em}
\end{table}

\paragraph{Using English tokens instead of special tokens for roles in templates.}
In Section~\ref{sec:method}, we mentioned that we use language-agnostic templates to facilitate the cross-lingual transfer. To further validate the effectiveness of the language-agnostic template. We conduct experiments using English tokens as the templates. Specifically, we set format
\\[7pt]
{\setlength{\tabcolsep}{1pt}
\begin{tabular}{|x{0.99\columnwidth}|}
    \hline
    \fontsize{7.8pt}{7.8pt}\selectfont
    Agent: \texttt{[None]} $<$SEP$>$ Victim: \texttt{[None]} $<$SEP$>$ Instrument: \\
    \fontsize{7.8pt}{7.8pt}\selectfont
    \texttt{[None]} $<$SEP$>$ Place: \texttt{[None]} \\
    \hline
\end{tabular}}
\\

\noindent to be the template for \emph{Life:Die} events. Hence, for non-English instances, the targeted output string is a code-switching sequence. Table~\ref{tab:enrole} lists the results. We can observe that applying language-agnostic templates bring \model{} 2.3 F1 scores improvements in average.

\begin{table}[t!]
\centering
\small
\setlength{\tabcolsep}{3.5pt}
\resizebox{\columnwidth}{!}{
\begin{tabular}{@{}l|ccc|ccc|c}
    \toprule
    Model
    &\makecell{en\\$\Downarrow$\\xx} & \makecell{ar\\$\Downarrow$\\xx} & \makecell{zh\\$\Downarrow$\\xx}
    &\makecell{xx\\$\Downarrow$\\en} & \makecell{xx\\$\Downarrow$\\ar} & \makecell{xx\\$\Downarrow$\\zh}
    & avg \\
    \midrule
    \model{} (mT5-base)
    & \textbf{54.3} & \textbf{41.4} & \textbf{51.4} & \textbf{49.4} & \textbf{46.7} & \textbf{51.0} & \textbf{49.1} \\
    \ w/ English Tokens
    & 51.4 & 39.3 & 49.7 & 46.6 & 44.7 & 49.0 & 46.8 \\
    \bottomrule
\end{tabular}}
\caption{Comparison of using English tokens and special tokens for roles in templates. ``en $\Rightarrow$ xx'' indicates the average of ``en $\Rightarrow$ en'', ``en $\Rightarrow$ zh'', and ``en $\Rightarrow$ ar''.}
\label{tab:enrole}
\vspace{-1em}
\end{table}

\subsection{Error Analysis}
\label{subsec:err}
We perform error analysis on \model{} (mT5-base) when transferring from Arabic to English and transferring from Chinese to English. For each case, we sample 30 failed examples and present the distribution of various error types in Figure~\ref{fig:err_analysis}.

\paragraph{Errors on both monolingual and cross-lingual models.} We compare the predicted results from \model{}(ar $\Rightarrow$ en) with \model{}(en $\Rightarrow$ en), or from \model{}(zh $\Rightarrow$ en) with \model{}(en $\Rightarrow$ en).
If their predictions are similar and both of them are wrong when compared to the gold output, we classify the error into this category. To overcome the errors in this category, the potential solution is to improve monolingual models for EAE tasks.

\paragraph{Over-generating.} 
Errors in this category happen more often in \model{}(ar $\Rightarrow$ en). It is likely because the entities in Arabic are usually much longer than that in English when measuring by the number of sub-words. Based on our statistics, the average entity span length is 2.85 for Arabic and is 2.00 for English (length of sub-words). This leads to the natural for our \model{}(ar $\Rightarrow$ en) to overly generate some tokens even though they have captured the correct concept.
An example is that the model predicts \textit{``The EU foreign ministers''}, while the ground truth is \textit{``ministers''}.

\paragraph{Label disagreement on different language splits.}
The annotations for the ACE dataset in different language split contain some ambiguity. For example, given sentence \textit{``He now also advocates letting in U.S. troops for a war against Iraq even though it is a fellow Muslim state.''} and the queried trigger \textit{``war''}, the annotations in English tends to label \textit{Iraq} as the \textit{Place} where the event happen, while similar situations in other languages will mark \textit{Iraq} as the \textit{Target} for the war.

\paragraph{Grammar difference between languages.}
An example for this category is \textit{``... Blackstone Group would buy Vivendi's theme park division, including Universal Studios Hollywood ...''} and the queried trigger \textit{``buy''}. 
We observe that \model{}(ar $\Rightarrow$ en) predicts \textit{Videndi} as the \textit{Artifact} been sold and \textit{division} is the \textit{Seller}, while \model{}(en $\Rightarrow$ en) can correctly understand that \textit{Videndi} are the \textit{Seller} and \textit{division} is the \textit{Artifact}. 
We hypothesize the reason being the differences between the grammar in Arabic and English.
The word order of the sentence~\textit{``Vivendi's theme park division''} in Arabic is reversed with its English counterpart, that is, \textit{``theme park division''} will be written before~\textit{``Vivendi''} in Arabic. Such difference leads to errors in this category.

\paragraph{Generating words not appearing in the passage.}

In \model{}(zh $\Rightarrow$ en), we observe several cases that generate words not appearing in the passage. There are two typical situations. The first case is that \model{}(zh $\Rightarrow$ en) mixes up singular and plural nouns. For example, the model generates~\textit{``studios''} as prediction while only \textit{``studio''} appears in the passage. This may be because Chinese does not have morphological inflection for plural nouns. 
The second case is that \model{}(zh $\Rightarrow$ en) will generate random predictions in Chinese.

\paragraph{Generating correct predictions but in Chinese.}
This is a special case of \textit{``Generating words not appearing in the passage''}. In this category, we observe that although the prediction is in Chinese (hence, a wrong prediction), it is correct if we translate the prediction into English.

\subsection{Constrained Decoding}

Among all the errors, we highlight two specific categories --- \textit{``Generating words not appearing in the passage''} and \textit{``Generating correct predictions but in Chinese''}. These errors can be resolved by applying constrained decoding~\cite{DBLP:journals/corr/abs-2103-12528} to force all the generated tokens to appear input.

Table~\ref{tab:cons_dec} presents the result of \model{} with constrained decoding. We observe that adapting such constraints indeed helps the cross-lingual transferability, yet it also hurts the performance in some monolingual cases. We conduct a qualitative inspection of the predictions. The observation is that constrained decoding algorithm although guarantees all generated tokens appearing in the input, the coercive method breaks the overall sequence distribution that learned. Hence, in many monolingual examples, once one of the tokens is corrected by constrained decoding, its following generated sequence changes a lot, while the original predicted suffixed sequence using beam decoding are actually correct. This leads to a performance decrease.\footnote{Indeed, a similar situation happens to cross-lingual cases; however, since the original performance for cross-lingual transfer is not high enough, the benefits of correcting tokens are more significant than this drawback.}


\begin{table}[t!]
\centering
\small
\setlength{\tabcolsep}{3.5pt}
\resizebox{\columnwidth}{!}{
\begin{tabular}{l|ccc}
    \toprule
    Model
    &\makecell{monolingual} & \makecell{cross-lingual} & \makecell{average all}
    \\
    \midrule
    \model{} (mBART-50-large)  
    & \textbf{63.9} & 37.4 & \textbf{46.2}   \\
    \ w/ constrained decoding 
    & 62.4 & \textbf{37.6} & 45.9 \\
    \midrule
    \model{} (mT5-base) 
    & \textbf{67.8} & 39.7 & \textbf{49.1}\\
    \ w/ constrained decoding 
    & 67.0 & \textbf{39.9} & 48.9\\
    \midrule
    \model{} (mT5-large) 
    & \textbf{69.7} & 42.2 & 51.3 \\
    \ w/ constrained decoding  
    & 68.8 & \textbf{43.1} & \textbf{51.6}\\
    \bottomrule

\end{tabular}}
\caption{Results of applying constrained decoding. Breakdown numbers can be found in Appendix~\ref{app:cons_dec}. Based on whether the training languages are the same between training and testing, we classify the results into \textit{monolingual} and \textit{cross-lingual}, and we report the corresponding average for each category.}
\vspace{-1em}
\label{tab:cons_dec}
\end{table}
\section{Conclusion}
We present the first generation-based models for zero-shot cross-lingual event argument extraction. To overcome the discrepancy between languages, we design language-agnostic templates and propose \model{}, which well capture output dependencies and can be used without additional named entity extraction modules.
Our experimental results show that \model{} outperforms the current state-of-the-art, which demonstrates the potential of using a language generation framework to solve zero-shot cross-lingual structured prediction tasks.

\section*{Acknowledgments}
We thank anonymous reviewers for their helpful feedback. We thank the UCLA PLUSLab and UCLA-NLP group for the valuable discussions and comments. We also thank Steven Fincke, Shantanu Agarwal, and Elizabeth Boschee for their help on data preparation in Arabic.
This work is supported in part by the Intelligence Advanced
Research Projects Activity (IARPA), via Contract
No.\ 2019-19051600007, and research awards sponsored by CISCO and Google.

\section*{Ethics Considerations}

Our proposed models are based on the multilingual pre-trained language model that is trained on a large text corpus. It is known that the pre-trained language model could capture the bias reflecting the training data. Therefore, our models can potentially generate offensive or biased content learned by the pre-trained language model. We suggest carefully examining the potential bias before deploying our model in any real-world applications.

\bibliography{acl}
\bibliographystyle{acl_natbib}

\clearpage
\appendix

\begin{table*}[t!]
\centering
\small
\setlength{\tabcolsep}{4pt}
\begin{tabular}{l|c|ccc|ccc|ccc}
    \toprule
    \multirow{2}{*}{Dataset}  & \multirow{2}{*}{Lang.} & 
    \multicolumn{3}{c|}{Train} & \multicolumn{3}{c|}{Dev} & \multicolumn{3}{c}{Test} \\
    & & \#Sent. & \#Event & \#Arg.  & \#Sent.& \#Event & \#Arg.  & \#Sent. & \#Event & \#Arg. \\
    \midrule
    \multirow{3}{*}{ACE-2005}
    & en & 17172& 4202 & 4859 & 923& 450 & 605 & 832& 403 & 576  \\
    & ar & 2722& 1743 & 2506 & 289& 117 & 174 & 272&  198 & 287  \\
    & zh & 6305& 2926 & 5581 & 486& 217 & 404 & 482& 190 & 336  \\
    \midrule
    \multirow{2}{*}{ERE}
    & en & 14734&6208 & 8924 & 1209&525 & 730 & 1161&551 & 882  \\
    & es & 4582&3131 & 4415 & 311&204 & 279 & 323&255 & 354  \\
    \bottomrule
\end{tabular}
\caption{Dataset statistics of ACE-2005 and ERE.}
\label{tab:data}
\end{table*}

\section{Dataset Statistics and Data Preprocessing}
\label{app:dataset}
Table~\ref{tab:data} presents the detailed statistics for the ACE-2005 dataset and ERE dataset.

For the English and Chinese splits in ACE-2005, we use the setting provided by \citet{Wadden19dygiepp} and \citet{Lin20oneie}, respectively. 
As for Arabic part, we adopt the setup proposed by \citet{Xu21arsplit}. Observing that part of the sentence breaks made from \citet{Xu21arsplit} being extremely long for pretrained models to encode, 
we perform additional preprocessing and postprocessing procedures for Arabic data. Specifically, we split Arabic sentences into several portions that any of the portion is shorter than 80 tokens. Then, we map the models' predictions of the split sentences back to the original sentence during postprocessing.

\section{Implementation Details}
\label{app:imp}
We describe the implementation details for all the models as follows:
\begin{itemize}[topsep=3pt, itemsep=1pt, leftmargin=12pt]
    \item \textbf{OneIE} \cite{Lin20oneie}. We use their provided code\footnote{\url{http://blender.cs.illinois.edu/software/oneie/}} to train the model with the provided default settings. It is worth mention that for the Arabic split in the ACE-2005 dataset, OneIE is trained with only entity extraction, event extraction, and event argument extraction since there is no relation labels in \citet{Xu21arsplit}'s preprocessing script. All other parameters are set to the default values.
    \item \textbf{CL-GCN} \cite{Subburathinam19clgcn}. We refer the released code from \citet{Wasi21gate}\footnote{\url{https://github.com/wasiahmad/GATE}} to re-implement the CL-GCN method. Specifically, we adapt the baseline framework that described and implemented in OneIE's code \cite{Lin20oneie}, but we remove its relation extraction module and add two layers of GCN on top of XLM-RoBERTa-large. The pos-tag and dependency parsing annotations are obtained by applying Stanza~\cite{Stanza}. All other parameters are set to the be the same as the training of OneIE.
    \item \textbf{GATE} \cite{Wasi21gate}. We refer the official released code from \citet{Wasi21gate} to re-implement GATE. Similar to CL-GCN, we adapt the baseline framework that described and implemented in OneIE's code, but we remove its relation extraction module and add two layers of GATE on top of XLM-RoBERTa-large, mT5, or mBART-50-large. The pos-tag and dependency parsing annotations are also obtained by applying Stanza~\cite{Stanza}. The hyper-parameter of $\delta$ in GATE is set to be [2, 2, 4, 4, $\infty$, $\infty$, $\infty$, $\infty$]. All other parameters are set to the be the same as the training of OneIE.
    \item \textbf{TANL} \cite{Paolini21tanl}. To adapt TANL to zero-shot cross-lingual EAE, we adapt the public code\footnote{\url{https://github.com/amazon-research/tanl}} and replace its pre-trained based model T5 \cite{t5model} with mT5-base \cite{Xue21mt5}. All other parameters are set to their default values.
    \item \textbf{\model{}} is our proposed model. We consider three different pre-trained generative language models: mBART-50-large \cite{mbart50}, mT5-base, and mT5-large \cite{Xue21mt5}. When fine-tune the pre-trained models, we set the learning rate to $10^{-4}$ for mT5, and $10^{-5}$ for mBART-50-large. The batch size is set to 8. The number of training epochs is 60.
\end{itemize}

\begin{table*}[t]
\centering
\small
\setlength{\tabcolsep}{5pt}
\resizebox{\textwidth}{!}{
\begin{tabular}{l|ccc|ccc|ccc|c|c|c}
    \toprule
    Model
    &\makecell{en\\$\Downarrow$\\en} & \makecell{en\\$\Downarrow$\\zh} & \makecell{en\\$\Downarrow$\\ar}
    &\makecell{ar\\$\Downarrow$\\ar} & \makecell{ar\\$\Downarrow$\\en} & \makecell{ar\\$\Downarrow$\\zh}
    &\makecell{zh\\$\Downarrow$\\zh} & \makecell{zh\\$\Downarrow$\\en} & \makecell{zh\\$\Downarrow$\\ar}
    &\makecell{avg \\ (mono.)} & \makecell{avg \\ (cross.)} & \makecell{avg \\ (all)}
    \\
    \midrule
    \model{} (mBART-50-large)  
    & \textbf{68.3} & 48.9 & 37.7 & \textbf{59.8} & 30.5 & 29.2 & \textbf{63.6} & 45.9 & \textbf{32.3} & \textbf{63.9} & 37.4 & \textbf{46.2}   \\
    \ w/ constrained decoding 
    & 68.0 & \textbf{49.1} & \textbf{37.8} & 59.5 & \textbf{30.6} & 29.2 & 59.7 & \textbf{47.7} & 31.3 & 62.4 & \textbf{37.6} & 45.9 \\
    \midrule
    \model{} (mT5-base) 
    & 67.9 & 53.1 & 42.0 & 66.2 & 27.6 & \textbf{30.5} & \textbf{69.4} & 52.8 & 32.0 & \textbf{67.8} & 39.7 & \textbf{49.1}\\
    \ w/ constrained decoding 
    & 67.9 & 53.1 & 42.0 & 66.2 & \textbf{27.8} & 30.4 & 66.7 & \textbf{53.1} & \textbf{33.1} & 67.0 & \textbf{39.9} & 48.9\\
    \midrule
    \model{} (mT5-large) 
    & 71.2 & 54.0 & 44.8 & 68.9 & \textbf{32.1} & 33.3 & \textbf{68.9} & 55.8 & 33.1 & \textbf{69.7} & 42.2 & 51.3 \\
    \ w/ constrained decoding  
    & 71.2 & \textbf{54.8} & \textbf{45.6} & 68.9 & 32.0 & 33.3 & 66.2 & \textbf{57.7} & \textbf{35.0} & 68.8 & \textbf{43.1} & \textbf{51.6}\\
    \bottomrule

\end{tabular}}
\caption{
The detailed breakdown results for applying constrained decoding on \model{}.
The avg(mono.) column represents the results that average over values in \textit{en $\Rightarrow$ en}, \textit{zh $\Rightarrow$ zh}, and \textit{ar $\Rightarrow$ ar}. The avg(cross.) column represents the results that average over values in \textit{en $\Rightarrow$ zh}, \textit{en $\Rightarrow$ ar}, \textit{zh $\Rightarrow$ en}, \textit{zh $\Rightarrow$ ar}, \textit{ar $\Rightarrow$ en}, and \textit{ar $\Rightarrow$ zh}.
}
\label{tab:detail_cons_dec}
\end{table*}

\section{Constrained Decoding Detailed Results}
\label{app:cons_dec}
Table~\ref{tab:detail_cons_dec} shows the detailed results for \model{} using constrained decoding algorithm during testing time. We directly apply constrained decoding algorithms on the trained models we have in Table~\ref{tab:ace}. 

\end{document}